\relax
\documentclass[letterpaper]{article}
\usepackage{times}
\usepackage{helvet}
\usepackage{courier}
\usepackage{amssymb}
\usepackage{amsmath}
\usepackage{mathtools}
\usepackage{bigdelim}
\usepackage{bbm}
\usepackage{mathtools}
\usepackage{centernot}
\usepackage{enumerate}
\usepackage{subcaption}
\usepackage{pgfplots}
\usepackage[caption=false,font=footnotesize]{subfig}
\usepackage{undertilde}
\usepackage{verbatim}
\usetikzlibrary{arrows,calc}
\usepackage{relsize}

\begin{document}
%
\title{Personalized Donor-Recipient Matching for Organ Transplantation}
\author{Jinsung Yoon, Ahmed M. Alaa, Martin Cadeiras, and Mihaela van der Schaar\\
University of California, Los Angeles\\
}
\maketitle
\begin{abstract}
Organ transplants can improve the life expectancy and quality of life for the recipient but carries the risk of serious post-operative complications, such as septic shock and organ rejection. The probability of a successful transplant depends in a very subtle fashion on compatibility between the donor and the recipient -- but current medical practice is short of domain knowledge regarding the complex nature of recipient-donor compatibility. Hence a data-driven approach for learning compatibility has the potential for significant improvements in match quality. This paper proposes a novel system (ConfidentMatch) that is trained using data from electronic health records. ConfidentMatch predicts the success of an organ transplant (in terms of the 3-year survival rates) on the basis of clinical and demographic traits of the donor and recipient. ConfidentMatch captures the heterogeneity of the donor and recipient traits by optimally dividing the feature space into clusters and constructing different optimal predictive models to each cluster. The system controls the complexity of the learned predictive model in a way that allows for assuring more granular and confident predictions for a larger number of potential recipient-donor pairs, thereby ensuring that predictions are ``personalized" and tailored to individual characteristics to the finest possible granularity. Experiments conducted on the UNOS heart transplant dataset show the superiority of the prognostic value of ConfidentMatch to other competing benchmarks; ConfidentMatch can provide predictions of success with 95$\%$ confidence for 5,489 patients of a total population of 9,620 patients, which corresponds to 410 more patients than the most competitive benchmark algorithm (DeepBoost). 
\end{abstract}

\section{Introduction}
Organ transplantation is the therapy of choice for patients with end stage diseases who are refractory to medical therapies (Shah 2012). Even though organ transplantation can increase the life expectancy and quality of life for the recipient, the operation can entail various complications, including infection, acute and chronic rejection and malignancy (Huynh 2014). Pre-operative anticipation of the risk associated with organ transplantation is a regular task that transplant centers perform in order to determine which patients would benefit from transplantation and accurately identify these for whom the risk of transplantation is too high and would therefore provide no survival benefits. Such a risk assessment task is quite complicated for that post-operative patient survival depends on different types of risk factors: recipient-related factors (e.g. cardiovascular disease severity of heart recipients (Wozniak 2014; Nwakanma 2007; Russo 2006; Silva 2016)), recipient-donor matching factors (e.g. weight ratio and HLA (Jayarajan 2013), blood group compatibility (Jawitz 2013), race (Allen 2010), etc), and donor-related factors (e.g. diabetes (Arnaoutakis 2012; Taghavi 2013)). The interactions among all these risk factors make the prognosis problem for the organ transplant outcomes highly complex; the NHLBI working group suggests resolving this problem by enhancing the phenotypic compatibility characterization of the pre-transplant recipient-donor population (Mancini 2010; Collins 2015; Shah 2012).   

In the light of the above, we seek an enhanced phenotypic characterization for the compatibility of patient-donor pairs via a {\it precision medicine} approach (Collins 2015) in which we construct {\it personalized} predictive models that are tailored to the individual traits of both the donor and the recipient to the finest possible granularity. The advent of electronic health records (EHR) inspire a data-driven approach for constructing such predictive models in which the complex recipient-donor compatibility patterns are discovered from observational data. To that end, we develop {\it ConfidentMatch}: an automated system that learns the recipient-donor compatibility patterns from the EHR data in terms of the probability of transplant success for given recipient-donor pairs. ConfidentMatch can be utilized by the clinicians as a prognostic tool for managing organ transplantation selection decisions in which information about the donor and recipient are fed to the system, and the output comes as the probability of the transplant's success. The system can also be used in conjunction with a matching algorithm, such as the Nobel prize winning algorithm of Shapley and Roth (Shapley 1962; Roth 2003), where a set of patients are matched to a set of donors given the compatibility scores that ConfidentMatch computes. 

In order to learn the highly complex recipient-donor compatibility patterns, ConfidentMatch adopts a novel learning framework in which the recipient-donor feature space is partitioned into disjoint subsets, and a separate predictive model (learner) is assigned to each partition. Such a learning approach gives rise to a highly complex overall predictive model for mapping recipient-donor features to transplant success probabilities. Over-fitting is controlled by penalizing the number of partitions in the recipient-donor feature space and the complexity of the learners assigned to the different partitions. Unlike existing meta-learning algorithms, ConfidentMatch solves an optimization problem through which it jointly determines how to partition the recipient-donor feature space, and what predictive model to assign to each partition. Since such an optimization problem is NP-hard, we propose an efficient greedy algorithm that proceed iteratively by first fixing the number of partitions in the recipient-donor feature space, optimizing the predictive models assigned to each partition, and then further stratifying each partition and re-assigning more ``specialized" predictive models to the new, finer partitions. The algorithm stops stratifying the recipient-donor feature space when further stratification would lead to new partitions with no enough per-partition training data for learning finer predictive models.  

Experiments conducted on the UNOS heart transplant dataset (Cecka 1996) show that ConfidentMatch can provide predictions of success with 95$\%$ confidence for 5,489 patients of a total population of 9,620 patients -- 410 more patients than for the best state-of-the-art machine learning algorithm (DeepBoost).

\section{Related Work}
We identify two broad categories of learning algorithms that are capable of combining multiple predictive models and/or stratifying the feature space into fine clusters: ensemble learning algorithms and clustering algorithms. We compare ConfidentMatch with these methods hereunder. \\
\\
{\bf Ensemble learning algorithms}\,\,\, Methods based on ensemble learning, such as Random Forest (Liaw 2002), LogitBoost (Friedman et al. 2000), Adaptive Boosting (Freund 1997) and DeepBoost (Kuznetsov 2014), operate by allocating different sets of training data to different weak learners, and then aggregating the predictions of these weak learners through a weighted sum to issue a final prediction (Dai et al. 2016) and (Tekin et al. 2016). While these methods can learn complex functions through the synergy of multiple weak learners, they do not integrate the allocation of the training data to the different learners (in both the bagging and boosting approaches) as part of their loss minimization problems. Therefore, these methods do not -- in principle -- learn a granular predictive model that performs well uniformly over the feature space, but rather learn a predictive model that works well ``on average". Contrarily, ConfidentMatch jointly optimizes the partitioning of the feature space together with the predictive model associated with each partition, and hence, it learns a refined recipient-donor phenotypic compatibility characterization in which the predictions are tailored to fine segments of the recipient-donor feature space, leading to an overall improved performance as compared to conventional ensemble methods. We will demonstrate the superiority of ConfidentMatch to ensemble learning algorithms in the ``Results and Discussion" Section.  \\  
\\
{\bf Clustering algorithms}\,\,\, Clustering is a natural approach for identifying phenotypic characterizations by grouping ``similar" patients (or recipient-donor pairs) into distinct clusters. Clustering algorithms can be divided into two categories: unsupervised and supervised clustering. Unsupervised clustering algorithms, such as the $k$-means algorithm (MacQueen 1967), utilize the feature space solely to learn the partitions that maximize some given objective, and hence they cannot address the organ transplant prognosis problem since they do not consider the transplant outcomes (i.e. labels) in the clustering process. 

Supervised clustering algorithms utilize both the feature space and the label space for constructing clusters (Eick 2004; Finley 2005); however, the predictive models assigned to each partition of the feature space are limited to indicator functions (an example for such algorithms is the regression tree (Strobl 2009)). For the organ transplant setting, the complex interactions between the donor and recipient features create highly complex patterns of recipient-donor compatibility (transplant success probabilities) that would exhibit very high per-partition impurity under conventional supervised clustering or tree learning algorithms. This means that learning such complex medical concepts would face the dilemma of exhibiting large over-fitting errors when adopting a large number of clusters (or very deep decision trees) to resolve the per-partition impurity, and exhibiting a large bias error when restricting the number of clusters (or restricting the depth of decision trees). ConfidentMatch approaches this problem by providing a versatile framework for complexity control where complex predictive models can be assigned to every partition to reduce the per-partition bias error, which enables learning complex functions with less partitions and hence utilize the training data more efficiently.   

\section{Methods}

Let the $D$-dimensional recipient-donor feature space be denoted as $\mathcal{X}$; every instance in $\mathcal{X}$ corresponds to a recipient-donor pair with certain given characteristics. Denote the corresponding label space which designates the success or failure of an organ transplant for a given recipient-donor pair as $\mathcal{Y}$; every label instance can be defined as the specific event of transplant success or failure if $\mathcal{Y} = \{0,1\}$, or the probability of the transplant's success if $\mathcal{Y} = [0,1]$. Let $\mathcal{T}$ be a dataset extracted from the EHR; we split the dataset $\mathcal{T}$ into two separate sets: a training set $\mathcal{S}=\{({\bf{x}_1^s},y_1^s),...,({\bf{x}_m^s},y_m^s)\}$ and a validation set $\mathcal{V}=\{({\bf{x}_1^v},y_1^v),...,({\bf{x}_n^v},y_n^v)\}$, where $\mathcal{S}$ and $\mathcal{V}$ are disjoint ($\mathcal{S} \cap \mathcal{V} = \emptyset$). Every entry in $\mathcal{S}$ and $\mathcal{V}$ comprise a recipient-donor feature pair and a transplant outcome. 

The goal of ConfidentMatch is to construct a predictive model $h \in \mathcal{H}, h: \mathcal{X} \rightarrow \mathcal{Y}$ that maps recipient-donor pairs to anticipated transplant outcomes; such a model has to be learned from the dataset $\mathcal{T} = \{\mathcal{S}, \mathcal{V}\}$, and can be used for out-of-sample recipient-donor pair in order to assess the risk of a recipient's transplant operation. The problem of learning the predictive model $h \in \mathcal{H}$ from the labeled dataset $\mathcal{T}$ is a standard supervised learning problem (Shalev-Shwartz 2014).  

The expected loss of a predictive model $h$ is defined as $\mathcal{L}_{\mathcal{F}}(h) =\mathbb{E}_{\mathcal{F}} [l(h({\bf{x}}),y)]$ where $l(h({\bf{x}}),y)$ is a general loss function, and $\mathcal{F}$ is the joint recipient-donor feature-label distribution, which is unknown to the clinicians. The optimal predictive model is defined as $h^* = \arg \min_{h \in \mathcal{H}} \mathcal{L}_{\mathcal{F}}(h)$; since $\mathcal{F}$ is unknown, we cannot directly find the optimal predictive model, and hence we resort to minimizing the empirical loss as measured over the training and validation sets. The empirical loss for the training set is defined as $\mathcal{L}_{\mathcal{S}}(h) = \cfrac{1}{m} \sum_{i=1}^m l(h({\bf{x_i^s}}),y_i^s)$, and it can be defined similarly for the validation set. 

Note that as pointed out in the previous section, the optimal predictive model $h^*$ is likely to be of a very complex structure as it abstracts a complex medical concept, i.e. the interactions between the recipient and donor features and their effect on the transplant outcomes. A poor initial choice for the space of possible models $\mathcal{H}$, e.g. letting $\mathcal{H}$ be a hypothesis class with a small VC dimension, may lead to a large bias in the loss function of $h^*$, and hence we need a more versatile learning framework for which the complexity of the predictive model $h$ adapts to the complexity of the underlying medical concept being learned.   

ConfidentMatch adopts a novel framework for crafting complex predictive models out of simpler baseline models by creating a phenotypic characterization of the recipient-donor feature space in which separate predictive models are assigned to disjoint partitions of the feature space. That is, ConfidentMatch outputs a set of partitions that cover the entire recipient-donor feature space, together with a set of predictive models, each tailored to a given partition, thereby leading to an overall complex, granular predictive model. Formally, we divide the recipient-donor feature space $\mathcal{X}$ into $k$ disjoint subsets, where $k$ is to be determined based on the given dataset, in such a way that for each subset, we can have a separate optimal predictive model that minimizes the overall expected risk. We write $\{\mathcal{X}_1,...,\mathcal{X}_k\}$ to denote a partition of the feature space $\mathcal{X}$, where all such partitions are ensured to be disjoint and cover $\mathcal{X}$. The partition $\{\mathcal{X}_1,...,\mathcal{X}_k\}$ can be translated to a partitioning of the training set $\mathcal{S}$ and validation set $\mathcal{V}$, i.e. the training set is partitioned as $\{\mathcal{S}_1,...,\mathcal{S}_k\}$, where ${\mathcal{S}}_i = \{({\bf{x}},y): {{(\bf{x}}, y) \in \mathcal{S} \mbox{ and } \bf{x}} \in \mathcal{X}_i\}$.

Given the above construct, the learning problem becomes a problem of (jointly) finding the optimal partitioning $\{\mathcal{X}_1,...,\mathcal{X}_k\}$ of the recipient-donor feature space, together with the optimal predictive model $h_i \in \mathcal{H}$ associated with every partition $i$, i.e. assuming that we know the distribution $\mathcal{F}$, the optimal predictive model is found by solving the following optimization problem  
\begin{equation}\label{eq1}
\begin{aligned}
& \underset{\{\mathcal{X}_1,...,\mathcal{X}_k\}}{\text{min}}
\Bigg[\underset{h_1,...,h_k \in \mathcal{H}}{\text{min}} & \sum_{i=1}^k \mathcal{F}(\boldsymbol{X} \in \mathcal{X}_i) \times \mathbb{E}_{\mathcal{F}_i} [l(h_i({\bf{x}}),y)] \Bigg] \\
& \text{subject to}
& \mathcal{X} = \bigcup_{i=1}^k \mathcal{X}_i \mbox{, and } \mathcal{X}_i \cap \mathcal{X}_j = \emptyset\; \forall i \neq j.
\end{aligned}
\end{equation}
We break down the problem in (\ref{eq1}) into two nested optimization problems; we first focus on the solution of the inner optimization problem and define its solution (for a given partitioning $\{\mathcal{X}_1,...,\mathcal{X}_k\}$) as  
\begin{align*}
d(\{\mathcal{X}_1,...,\mathcal{X}_k\}) &= \underset{h_1,...,h_k \in \mathcal{H}}{\text{min}}  \sum_{i=1}^k \mathcal{F}(\boldsymbol{X} \in \mathcal{X}_i) \times \mathbb{E}_{\mathcal{F}_i} [l(h_i({\bf{x}}),y)]. 
\end{align*}
Note that given a partition $\{ \mathcal{X}_1,...,\mathcal{X}_k \}$, the solutions to the inner optimizations are separable, i.e. the optimal predictor of one partition can be determined independent of the choice of the predictors for the other partitions. Hence, we can simplify the inner optimization problem as follows.
\begin{align*}
\underset{h_1,...,h_k \in \mathcal{H}}{\text{min}}  &\sum_{i=1}^k \mathcal{F}(\boldsymbol{X} \in \mathcal{X}_i) \times \mathbb{E}_{\mathcal{F}_i} [l(h_i({\bf{x}}),y)]  = \nonumber \\
&\sum_{i=1}^k \min_{h_{i} \in \mathcal{H}} \mathcal{F}(\boldsymbol{X} \in \mathcal{X}_i) \times  \mathbb{E}_{\mathcal{F}_i} [l(h_i({\bf{x}}),y)].
\end{align*}

Since ConfidentMatch has no access to the true distribution $\mathcal{F}$, the algorithm has to learn the partitioning $\{ \mathcal{X}_1,...,\mathcal{X}_k \}$ and the corresponding predictive models $\{h_i\}_{i=1}^{k}$ from the dataset $\mathcal{T} = \mathcal{S} \cup \mathcal{V}$) in such a way that it reaches a loss function that is as close as possible to the true loss in (\ref{eq1}). To achieve this, we construct a proxy for the objective in (\ref{eq1}) by replacing the terms $\mathcal{F}(\boldsymbol{X} \in \mathcal{X}_i)$ and $\mathbb{E}_{\mathcal{F}_i} [l(h_i({\bf{x}}),y)]$, which depend on the unknown $\mathcal{F}$, with their sample estimates $\cfrac{|\mathcal{V}_i|}{n}$ and $\mathcal{L}_{\mathcal{V}_i}(h_i)= \cfrac{1}{n} \sum_{i=1}^n l(h_i({\bf{x_i^v}}),y_i^v)$. Hence, empirical loss minimization over the validation dataset $\mathcal{V}$ can be formulated as follows
\begin{equation}
\begin{aligned}
& \underset{k,\mathcal{X}_1,...,\mathcal{X}_k}{\text{min}}
\Bigg[ \sum_{i=1}^k   &  \underset{h_i \in \mathcal{H}}{\text{min}} \cfrac{|\mathcal{V}_i|}{n} \times \cfrac{1}{|\mathcal{V}_i|} \sum_{({\bf{x_j^v}},y_j^v) \in \mathcal{V}_i} l(h_i({\bf{x_j^v}}),y_j^v) \Bigg] \\
& \text{subject to}
& \mathcal{X} = \bigcup_{i=1}^k \mathcal{X}_i \mbox{, and } \mathcal{X}_i \cap \mathcal{X}_j = \emptyset \mbox{ for } \forall i \neq j.
\label{eq2}
\end{aligned}
\end{equation}
ConfidentMatch constructs the hypothesis class $\mathcal{H}$ in (\ref{eq2}), from which we select a hypothesis (or a predictive model) $h_i$ for every partition $i$, by combining the outputs of a finite set of $M$ learners $\{\mathcal{A}_1,...,\mathcal{A}_M\}$, each of which can learn a predictive model that belongs to some hypothesis class $\mathcal{H}(\mathcal{A})$. That is, for a fixed partition $\mathcal{X}_i$, using the corresponding training set $\mathcal{S}_i$, the predictive model that is learned by the learning algorithm $\mathcal{A}_i$ for partition $i$ is $\mathcal{A}_{i}(\mathcal{S}_i)$. Therefore, the set of all predictive models that can be learned by all the learning algorithms operating on data set $\mathcal{S}_i$ is given as $\hat{\mathcal{H}}(\mathcal{S}_i)= \{ \mathcal{A}_1(\mathcal{S}_i),...,\mathcal{A}_M(\mathcal{S}_i) \}$. ConfidentMatch decides the optimal partitioning and the optimal predictor for each partition $i$ that belongs to a set of learnable predictors $\mathcal{\hat{H}}(\mathcal{S}_{i})$ by minimizing the empirical loss with respect to the validation data set as follows
\begin{equation}
\begin{aligned}
& \underset{k,\mathcal{X}_1,...,\mathcal{X}_k}{\text{min}}
\Bigg[ \sum_{i=1}^k   &  \underset{h_i \in \hat{\mathcal{H}}(\mathcal{S}_{i})}{\text{min}} \cfrac{|\mathcal{V}_i|}{n} \times \cfrac{1}{|\mathcal{V}_i|} \sum_{({\bf{x_j^v}},y_j^v) \in \mathcal{V}_i} l(h_i({\bf{x_j^v}}),y_j^v) \Bigg] \\
& \text{subject to}
& \mathcal{X} = \bigcup_{i=1}^k \mathcal{X}_i \mbox{,and } \mathcal{X}_i \cap \mathcal{X}_j = \emptyset \mbox{ for } \forall i \neq j.
\label{eq3}
\end{aligned}
\end{equation}

Note that the formulation in (\ref{eq3}) does not account for the out-of-sample error (or over-fitting); to handle that, we reformulate problem (\ref{eq3}) by replacing the objective function with a tight upper bound on the true loss that appropriately penalizes over-fitting. Define  
\begin{align}
\hat{d}(\{\mathcal{X}_{1},..\mathcal{X}_{k}\}) &= \sum_{i=1}^k \min_{h_i \in \hat{\mathcal{H}}(\mathcal{S}_i)} \Bigg[\cfrac{1}{n} \sum_{({\bf{x_j^v}},y_j^v) \in \mathcal{V}_i} l(h_i({\bf{x_j^v}}),y_j^v)\Bigg] \nonumber \\    
&+ \alpha \sqrt{\cfrac{k^2 \log{M}}{n}},
\label{eq4}
\end{align} 
where $\alpha \geq 0$ is a {\it penalty parameter}. The expression in (\ref{eq4}) comprises the sample estimate of the objective and a penalty term $\alpha \sqrt{\cfrac{k \log{M}}{n/k}}$ that penalizes: the number of partitions $k$, the average size of a partition $n/k$, and the number of predictive model $M$ from which we chose one model to assign to a given partition. It can be shown that if the penalty parameter $\alpha \geq \sqrt{\cfrac{1}{2} + \cfrac{1}{2 \log M} \log(\cfrac{2}{1-(1-\delta)^{1/k}}) }$, then the probability that $d(\{\mathcal{X}_{1},..\mathcal{X}_{k}\})$ is bounded above by $\hat{d}(\{\mathcal{X}_{1},..\mathcal{X}_{k}\})$ is greater than $1-\delta$, i.e. $\mathbb{P}( d(\{ \mathcal{X}_1,...,\mathcal{X}_k \}) < \hat{d}(\{ \mathcal{X}_1,...,\mathcal{X}_k \})  ) \geq 1-\delta$ (the proof can be found in the supporting material). By using the upper bound $\hat{d}(\{\mathcal{X}_{1},..\mathcal{X}_{k}\})$ as the objective, the empirical loss minimization problem becomes
\begin{equation}
\begin{aligned}
& \underset{\{\mathcal{X}_1,...,\mathcal{X}_k\}}{\text{min}}
&\Bigg[\sum_{i=1}^k  \underset{h_i \in \hat{\mathcal{H}(\mathcal{S}_i)}}{\text{min}}\cfrac{1}{n} \sum_{({\bf{x}_j^v},y_j^v) \in \mathcal{V}_i} l(h({\bf{x}_j^v}),y_j^v) \Bigg]\\ 
&+ \alpha \sqrt{\cfrac{k^2 \log{M}}{n}} \\
& \text{subject to}
& \mathcal{X} = \bigcup_{i=1}^k \mathcal{X}_i \mbox{, and } \mathcal{X}_i \cap \mathcal{X}_j = \emptyset \mbox{ for } \forall i \neq j
\label{eq5}
\end{aligned}
\end{equation}
Solving (\ref{eq5}) is computationally intractable because the number of possible partitions for $n$ points grows as the Bell number. To address this problem, ConfidentMatch adopts an efficient greedy algorithm for approximating the solution to (\ref{eq5}). As a first step to construct such an algorithm, we reformulate (\ref{eq5}) by incorporating two more constraints. First, we restrict the partitions of the recipient-donor feature space to be hypercubes. A hypercubic partition of the feature space $\mathcal{X}$ is defined as $\{\mathcal{X}_{1},..,\mathcal{X}_{k}\}$ where $\mathcal{X}_i = \prod_{j=1}^D [a_{ij},b_{ij}], a_{ij}\leq b_{ij}, a_{ij} \in \mathbb{R}^*, b_{ij} \in \mathbb{R}^*$. ($\mathbb{R}^*$ is defined as $\mathbb{R} \cup\{-\infty, \infty\}$). Second, we restrict the number of partitions to be $\gamma \in \mathbb{N}$. The optimization problem in (\ref{eq5}) with these additional constraints can be stated as follows.
\begin{equation}
\begin{aligned}
& \underset{\{\mathcal{X}_1,...,\mathcal{X}_k\}}{\text{min}}
& \Bigg[ &\sum_{i=1}^k  \underset{h_i \in \hat{\mathcal{H}}(\mathcal{S}_i)}{\text{min}} \cfrac{1}{n} \sum_{({\bf{x}_j^v},y_j^v) \in \mathcal{V}_i} l(h({\bf{x}_j^v}),y_j^v) \Bigg] \\
& & & + \alpha \sqrt{\cfrac{k^2 \log{M}}{n}}  \\
& \text{subject to}
& & \mathcal{X}_i = \prod_{j=1}^D [a_{ij},b_{ij}], a_{ij}\leq b_{ij}, a_{ij} \in \mathbb{R}^*, b_{ij} \in \mathbb{R}^* \\
& & & k \leq \gamma, \mbox{ where } k \in \mathbb{Z}_{+} \\
&
& & \mathcal{X} = \bigcup_{i=1}^k \mathcal{X}_i \mbox{, and } \mathcal{X}_i \cap \mathcal{X}_j = \emptyset \mbox{ for } \forall i \neq j.
\label{eq6}
\end{aligned}
\end{equation}

Let $Opt(\mathcal{X})$ be the optimal partition that solves the optimization problem in (\ref{eq6}); we construct a greedy algorithm which iteratively solve the optimization problem (\ref{eq6}) to achieve the approximate solution for (\ref{eq5}) as follows. We write the partition that is generated when $\mathcal{X}$ is input to the optimization problem (\ref{eq6}) as $Opt(\mathcal{X})=\{\mathcal{X}_{1},\mathcal{X}_{2},..,\mathcal{X}_{k}\}$ where $k < \gamma$. We apply the same procedure recursively on each $\mathcal{X}_{i}$ separately up to the point where we do not expect to improve the objective function. The final partition is the union of the partitions that are generated by applying this procedure recursively to each $\mathcal{X}_{i}$. We write the final partition achieved by the greedy algorithm as $\{ \hat{\mathcal{X}}_1^*,...,\hat{\mathcal{X}}_k^* \}$. 

The pseudo-code for ConfidentMatch is given in Fig. \ref{fig:pseudo}. The algorithm learns the recipient-donor compatibility patterns in an offline manner using the procedure described above: it jointly optimizes the partitioning of the recipient-donor feature space (offline stage I), and then optimizes the predictive model associated with each partition (offline stage II). Having learned the recipient-donor compatibilities, the algorithm operates in an online stage for new recipients and donors by computing a {\it compatibility score}, i.e. the probability of transplant success, for a given recipient-donor pair. The compatibility score is displayed to the clinicians and based on it, the clinicians/patients can make decisions on whether a transplant should be conducted, or whether the recipient should be matched with another donor.

In what follows, we specify the computational complexity of ConfidentMatch. Let the complexity of the algorithm $A_l(S_i)$ for learning a predictive model using the dataset $\mathcal{S}_i$ be $T_l(|S_i|,D)$. Based on this, it can be shown that the worst-case complexity for computing $Opt(\mathcal{X})$ in the optimization problem (\ref{eq6}) is $\mathcal{O}(\gamma^{\gamma+1} n^{\gamma} D^{\gamma-1} \sum_{i=1}^M T_i(n,D))$, and the complexity of the greedy algorithm described in Fig. \ref{fig:pseudo} is $\mathcal{O}(\gamma^{\gamma+1} n^{\gamma+1} D^{\gamma-1} \sum_{i=1}^M T_i(n,D))$ (proofs are provided in the supporting material).
	
\section{Results and Discussion}
Experiments were conducted using the UNOS database for patients who underwent a heart transplant over the years from 1987 to 2015 (Cecka 1996). We use the "Thoracic DATA" dataset in the UNOS database as our root dataset. In this dataset, all patients were followed-up until death, i.e. the post-transplant survival times for all patients are available in the dataset. Of the 148,512 patients in the “Thoracic DATA” who underwent either heart or lung transplant, we extract 60,516 patients who underwent a heart transplant. Of the 60,516 patients who underwent a heart transplant, we exclude 3,800 patients who are still alive, and we only use the 56,716 patients for whom we have the exact survival (lifetime) information. 

For each patient in the dataset, a total of 504 features are provided; these include a combination of both the patient's and the donor's information. We discard 12 features that are normally obtained after the transplant. Of the rest 492 features, we extract 70 features for which we have less than 10$\%$ missing information in order to reduce the noise of imputation. We use the KNN imputation method to impute the missing data (Hastie 1999). 

We compared the performance of ConfidentMatch in predicting the success of transplants with the following benchmark algorithms: logistic regression (Logit), Lasso regularized logistic re-gression (Lasso), decision tree (DTree), Random Forest (RForest), AdaBoost (ABoost), and DeepBoost (DBoost). We use the correlation feature selection (CFS) method to discover the relevant features for both ConfidentMatch and the benchmark algorithms (Hall 1999). We adopt the following metric for quantifying the performance of the different algorithms. The transplant's success probability is quantified via the 3-year post-transplant survival rate (long-term survival rate). We say that an algorithm provides a prediction for the transplant's success (probability of 3-year post-transplant survival) for a certain recipient-donor pair with a {\it confidence level $X\%$} if the algorithm's probability of correct prediction is $X\%$ for that recipient-donor pair. Based on this definition, we define the {\it gain} of ConfidentMatch at a confidence level of $X\%$ as the number of recipient-donor instances in the testing dataset for which ConfidentMatch provides a confident prediction for the transplant success, whereas the best competing benchmark does not. 

We split the dataset into a training set comprising the recipient-donor instances in which the recipient underwent the transplant before the year 2010 (past patients), and a testing set that comprises recipients who underwent the transplant after 2010 (current patients). Of the 56,716 recipient-donor pairs, 47,096 pairs (83.04$\%$) were used for training, and 9,620 pairs (16.96$\%$) were used for testing. We varied the confidence level from 80$\%$ to 95$\%$ and evaluated the performance within this range of confidence levels. 

Table 1 and Fig 2 show that ConfidentMatch consistently outperforms the benchmarks to predict the success of heart transplant regarding 3-year mortality prediction. For instance, ConfidentMatch boosts the number of recipient-donor pairs who get 95$\%$ confident predictions by 410 as compared to the best performing benchmark (DBoost); this means that ConfidentMatch can allow an additional number of 410 recipient-donor pairs to make better pre-transplant decisions such as whether or not the transplant should be conducted, and whether the recipient should be matched with another donor.

\begin{table}[h!]
	\begin{small}
		\begin{center}
			\begin{tabular}{|c|c|c|c|c|c|c|}
				\hline \hline
				\textbf{Conf.$^{*}$} & \textbf{CM$^{*}$} & \textbf{LASSO} & \textbf{RF$^{*}$}  & \textbf{AB$^{*}$} & \textbf{DB$^{*}$} & \textbf{DTree}
				\\ 
				\hline \hline	
				80\% & 9269	&	7721&	9057&	9063&	9067&	7893\\ \cline{1-7}
				85\% &7798	&	7096&	7605&	7577&	7614&	7316
				\\ \cline{1-7} 
				90\% & 6467	& 4431& 6065&	6076&	6080&	4526				
				\\ \cline{1-7} 
				95\% & 5489	& 2796&	5065&	5058&	5079&	3124
				\\ \hline	\hline
			\end{tabular}
		\end{center}
	\end{small}
	\label{tab:MResult1}
	\captionsetup{font= small}
	\caption{Confident predictions for the success of heart transplant by different algorithms ($^{*}$ Conf. = confidence level, CM = ConfidentMatch, RF = random forest, AB = ABoost, DB = DBoost).}
\end{table}	

\begin{table*}[h!]
	\begin{center}
		\begin{small}
			\begin{tabular}{|c|c|c|c|}
				\hline \hline
				& \multicolumn{3}{|c|}{\textbf{Relevant Features}} \\
				\hline					
				\textbf{No} & \textbf{Overall recipient-donor population} & \textbf{Group A} & \textbf{Group B} \\
				\hline
				1 & Ventilator assist at trr&	Ventilator assist at trr&	Ecmo assist at trr\\
				\hline
				2 &Ecmo assist at trr&	Ecmo assist at trr	&Ventilator assist at trr			 \\
				\hline
				3 & Other life support&	Other vent support at trr&	Donor VDRL Result \\
				\hline
				4 & Other vent support at trr&	Other life support&	Days in state 1A	 \\
				\hline
				5 &Days in state 1A	&Life support at trr&	Prior cardiac surgery\\
				\hline
				6 & Diagnosis&	Diagnosis&	Blood type matching\\
				\hline
				7 & Donor status (Decease)&	Donor status (Decease)&	Donor Blood Type (O)  \\
				\hline
				8 & Transplant type&	Malignancy&	Donor status (Decease)  \\
				\hline 
				9 & Donor HEP P surface antigen&	Transplant type	&Donor HEP B surface antigen \\
				\hline
				10 & History of previous MI	&Number of previous tx&	Inhaled assist  \\
				\hline \hline
			\end{tabular}
		\end{small}
	\end{center}
	\label{tab:MRel1}
	\captionsetup{font= small}
	\caption{Top 10 relevant features for the prediction of heart transplant success.}
\end{table*}

The performance gains achieved by ConfidentMatch can be attributed to the improved phenotypic characterization of the recipient-donor pairs that the algorithm achieves by stratifying the recipient-donor feature space. The fine and granular phenotypic characterization achieved by ConfidentMatch is restricted by the size of the training data; the more recipient-donor instances are available in the training set, the larger is the number of partitions that ConfidentMatch can construct and cast a specialized predictive model to. Fig 3 illustrates the trade-off associated with increasing the complexity of ConfidentMatch's predictive model by increasing the number of partitions; if the number of partitions increases, the gain of ConfidentMatch also increases as it copes with the underlying complexity of the recipient-donor compatibility patterns, until a certain number of partitions when the gain starts to decrease due to over-fitting.

\begin{figure}[t]
    \centering
    \includegraphics[width=3.25 in]{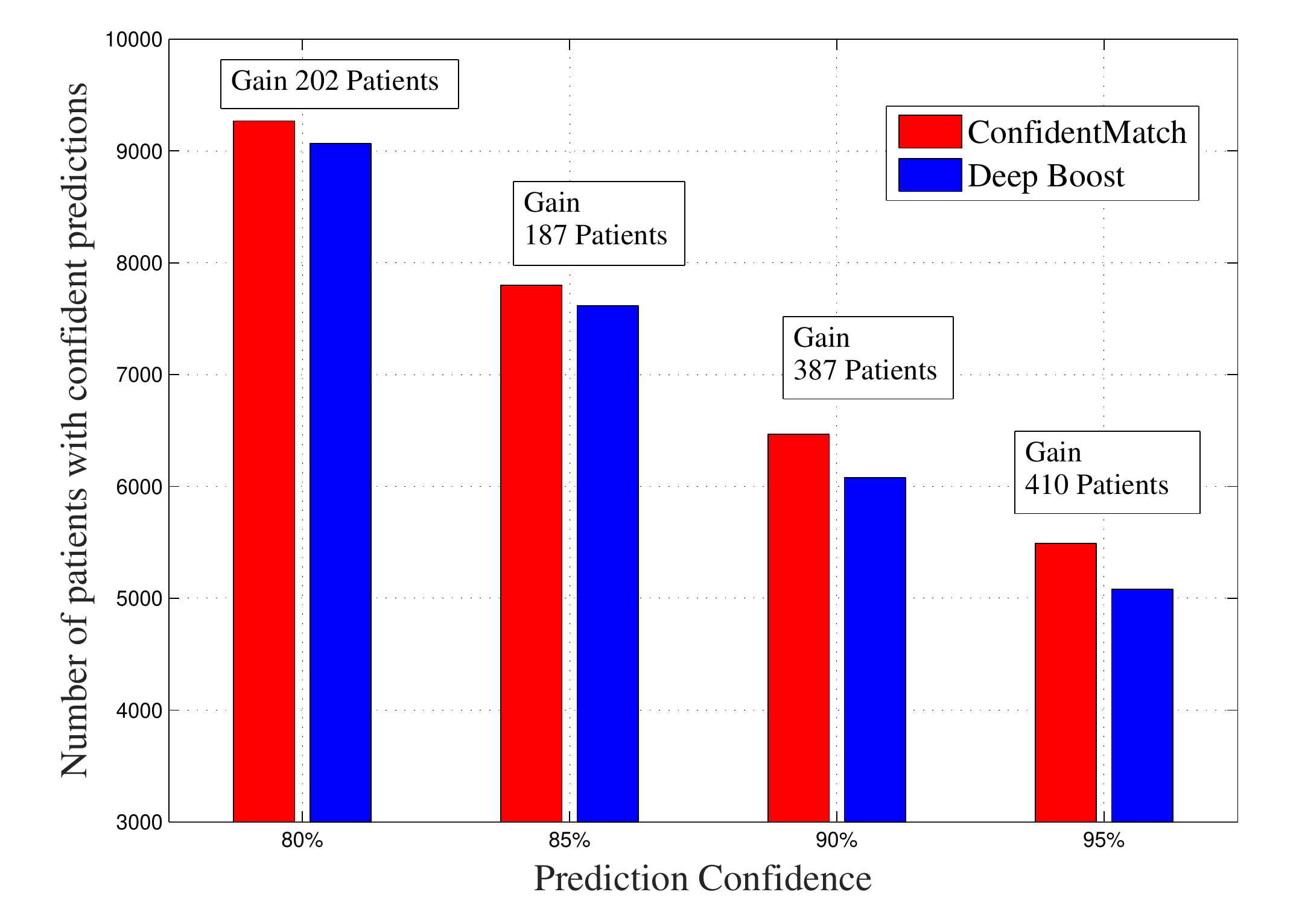}   
		\captionsetup{font= small}
    \caption{The number of recipient-donor pairs with confident predicitions based on ConfidentMatch as compared to DBoost.}
		\label{Fg22x}
\end{figure}

\begin{figure}[t]
    \centering
    \includegraphics[width=3.25 in]{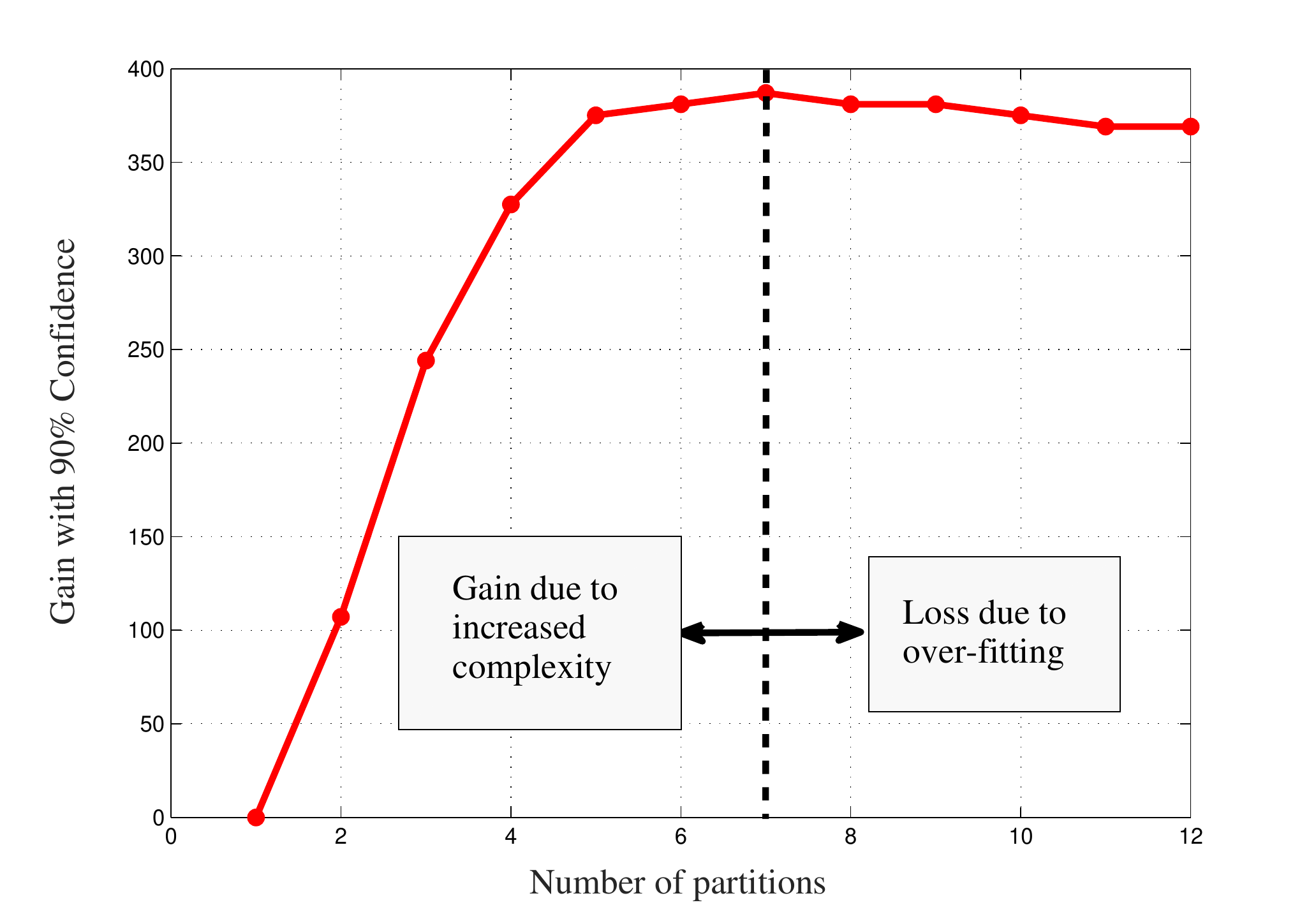}   
		\captionsetup{font= small}
    \caption{Trade-offs associated with increasing the number of partitions in ConfidentMatch (gain is with respect to DBoost).}
		\label{Fg2}
\end{figure}

ConfidentMatch does not only improve the quality of prognosis, but can also draw clinical insights on the patterns of recipient-donor compatibility. To illustrate this, we list the first two partitions through which ConfidentMatch stratifies the recipient feature space: ConfidentMatch forms two recipient groups, group A comprises patients whose length of stay in status 1A (urgent transplant wait-list) is shorter than 10 days, whereas group B comprises the remaining patients. Table 2 lists the most relevant features that are predictive of the transplant outcome for each group. It can be seen that group B patients are more sensitive to the donor characteristics; the donor's VDRL result and blood type are relevant to the transplant outcome, whereas group A patient appear to be less sensitive to these features. Thus, the more the patient waits in status 1A, the more it becomes essential to consider the extent of her compatibility with the donors. As more training data becomes available, ConfidentMatch can reveal finer partitions and identify the relevant features for more granular recipient subgroups. 

\section{Conclusions}
Organ transplants for patients with end stage diseases carries the risk of various serious post-operative complications, pre-operative anticipation of the transplant outcome depends on the compatibility between the donor and the recipient. In this paper, we have developed ConfidentMatch, a data-driven system that learns complex recipient-donor compatibility patterns from the outcomes of previous transplants. ConfidentMatch captures the complexity of such compatibility patterns by optimally dividing the recipient-donor feature space into clusters and assigning different optimal predictive models to each cluster, thereby ensuring that predictions are ``personalized" and tailored to individual characteristics of both the donors and the recipients. Experiments conducted on a public heart transplant dataset demonstrate the superiority of ConfidentMatch to other competing benchmark algorithms.

\newpage
\section{References}
\smallskip \noindent Kause, J. et al. 2004. A comparison of antecedents to cardiac arrests, deaths and emergency intensive care admissions in Australia and New Zealand, and the United Kingdom—the ACADEMIA study. \textit{Resuscitation} 62: 275--282. 

\smallskip \noindent Mancini D., Lietz K. 2010. Selection of cardiac transplantation candidates in 2010. \textit{Circulation} 122: 173--83. 
 
\smallskip \noindent Huynh, T. N., Kleerup, E. C., Raj, P. P., and Wenger, N. S. 2014. The opportunity cost of futile treatment in the intensive care unit. \textit{Critical Care Medicine}, 42: 1977--1982. 
 
\smallskip \noindent Shah M. R., Starling R. C., Schwartz L. L., Mehra M. R. 2012. Heart transplantation research in the next decade--a goal to achieving evidence-based outcomes: National Heart, Lung, And Blood Institute Working Group. \textit{J Am Coll Cardiol}, 59: 1263--1269. 
 
\smallskip \noindent Collins F. S., Varmus H. 2015. A new initiative on precision medicine. \textit{New England Journal of Medicine}. 372: 793--795. 
 
\smallskip \noindent Nilsson J., Ohlsson M., Höglund P., Ekmehag B., Koul B., and Andersson B. 2015. The International Heart Transplant Survival Algorithm (IHTSA): a new model to improve organ sharing and survival. \textit{PLoS One}.  

\smallskip \noindent Wozniak C. J., Stehlik J., Baird B. C., et al. 2014. Ventricular Assist Devices or Inotropic Agents in Status 1A Patients? Survival Analysis of the United Network of Organ Sharing Database. \textit{The Annals of thoracic surgery} 97: 1364--1372. 

\smallskip \noindent Nwakanma L. U., Williams J. A., Weiss E. S., Russell S. D., Baumgartner W. A., Conte J. V. 2007. Influence of pretransplant panel-reactive antibody on outcomes in 8,160 heart transplant recipients in recent era. \textit{The Annals of Thoracic Surgery} 84: 1556--1562. 
 
\smallskip \noindent Russo M. J., et al. 2006. Survival after heart transplantation is not diminished among recipients with uncomplicated diabetes mellitus: an analysis of the United Network of Organ Sharing database. \textit{Circulation}. 114: 2280--2287. 
 
\smallskip \noindent Enciso, J. S. et al. 2014. Effect of Peripheral Vascular Disease on Mortality in Cardiac Transplant Recipients (from the United Network of Organ Sharing Database). \textit{The American journal of cardiology} 114: 1111--1115. 
 
\smallskip \noindent Jayarajan S. N., Taghavi S., Komaroff E., Mangi A. A. 2013.  Impact of low donor to recipient weight ratios on cardiac transplantation. \textit{Journal on Thoracic Cardiovascular Surgery} 146: 1538--1543. 

\smallskip \noindent Jawitz O. K., Jawitz N., Yuh D. D., Bonde P. 2013. Impact of ABO compatibility on outcomes after heart transplantation in a national cohort during the past decade. \textit{Journal on Thoracic Cardiovascular Surgery} 146: 1239-1245.

\smallskip \noindent Allen J. G., Weiss E. S., Arnaoutakis G. J., Russell S. D., Baumgartner W. A., Conte J. V., and Shah A. S. 2010. The impact of race on survival after heart transplantation: an analysis of more than 20,000 patients. \textit{The Annals of thoracic surgery} 89:1956--1963.
 
\smallskip \noindent Taghavi S., Jayarajan S. N., Wilson L. M., Komaroff E., Testani J. M., Mangi A. A. 2013. Cardiac transplantation can be safely performed using selected diabetic donors. \textit{Journal on Thoracic Cardiovascular Surgery} 146: 442--447. 
 
\smallskip \noindent Arnaoutakis G. J., George T. J., Allen J. G., Russell S. D., Shah A. S., Conte J. V. and Weiss E. S. 2012 Institutional volume and the effect of recipient risk on short-term mortality after orthotopic heart transplant. \textit{Journal on Thoracic Cardiovascular Surgery} 143: 157--167. 

\smallskip \noindent Gale, D., Shapley, L. S. 1962. College Admissions and the Stability of Marriage. \textit{American Mathematical Monthly} 69: 9--14.

\smallskip \noindent Liaw, A., Wiener, M. 2002. Classiﬁcation and regression by random-forest. \textit{R news} 3: 18--22.

\smallskip \noindent Friedman, J., Hastie, T., Tibshirani, R. et al. 2000. Additive logistic regression: a statistical view of boosting. \textit{The annals of statistics} 28: 337--407.

\smallskip \noindent Freund, Y., Schapire, R. E. 1997. A decision-theoretic generalization of on-line learning and an application to boosting. \textit{Journal of computer and system sciences} 55: 119--139.

\smallskip \noindent Kuznetsov, V., Mohri, M., Syed, U. 2014. Multi-class deep boosting. \textit{In NIPS}, 2501--2509.

\smallskip \noindent MacQueen, J. et al. 1967. Some methods for classiﬁcation and analysis of multivariate observations. \textit{In Proceedings of the ﬁfth Berkeley symposium on mathematical statistics and probability} 1: 281--297.

\smallskip \noindent Eick, C. F., Zeidat, N., Zhao, Z. 2004. Supervised clustering-algorithms and beneﬁts. \textit{In IEEE ICTAI} 774--776.

\smallskip \noindent Finley, T., Joachims, T. 2005. Supervised clustering with support vector machines. \textit{In ICML} 217--224. 

\smallskip \noindent Strobl, C., Malley, J., Tutz, G. 2009. An introduction to recur-sive partitioning: rationale, application, and characteristics of classiﬁcation and regression trees, bagging, and random forests. \textit{Psychological methods} 14: 323.

\smallskip \noindent Cecka J. M. 1996. The unos scientiﬁc renal transplant registry–ten years of kidney transplants. \textit{Clinical transplants} 1–14.

\smallskip \noindent Hastie, T., Tibshirani, R., Sherlock, G., Eisen, M., Brown, P., and Botstein D. 1999, Imputing missing data for gene expression arrays.

\smallskip \noindent Hall, M. A. 1999. Correlation-based feature selection for machine learning. PhD thesis, \textit{The University of Waikato}.

\smallskip \noindent Shalev-Shwartz, S., Ben-David, S. 2014. Understanding machine learning: From theory to algorithms. \textit{Cambridge University Press}.

\smallskip \noindent Roth, A. E., Sonmez, T., Unver, M. U. 2003. Kidney exchange. \textit{National Bureau of Economic Research}.

\smallskip \noindent Dai, P., Gwadry-Sridhar, F., Bauer, M., Borrie, M. 2016. Bagging Ensembles for the Diagnosis and Prognostication of Alzheimer's Disease. \textit{Thirtieth AAAI Conference on Artificial Intelligence}.

\smallskip \noindent Tekin, C., Yoon, J., van der Schaar, M. 2016. Adaptive ensemble learning with confidence bounds for personalized diagnosis. \textit{AAAI Workshop on Expanding the Boundaries of Health Informatics using AI (HIAI’16): Making Proactive, Personalized, and Participatory Medicine A Reality}.

\end{document}